 \documentclass[pmlr,twocolumn]{jmlr} 
\let\SUP\textsuperscript
 \usepackage{rotating}

\usepackage{booktabs}
\usepackage[load-configurations=version-1]{siunitx} 

\theorembodyfont{\upshape}
\theoremheaderfont{\scshape}
\theorempostheader{:}
\theoremsep{\newline}

\jmlrvolume{ML4H Extended Abstract Arxiv Index}
\jmlryear{2020}
\jmlrsubmitted{2020}
\jmlrpublished{}
\jmlrworkshop{Machine Learning for Health (ML4H) 2020} 

\title[NLP to Detect Cognitive Concerns]{Natural Language Processing to Detect Cognitive Concerns in Electronic Health Records Using Deep Learning}

\author{Zhuoqiao Hong\nametag{\thanks{Authors contributed equally}\SUP{1}},
Colin G. Magdamo\nametag{\footnotemark[1]\SUP{1}}, 
Yi-han Sheu\nametag{\footnotemark[1]\SUP{1}},
Prathamesh Mohite\SUP{1},
Ayush Noori\SUP{1},
Elissa M. Ye\SUP{1},
Wendong Ge\SUP{1},
Haoqi Sun\SUP{1},
Laura Brenner\SUP{1},
Gregory Robbins\SUP{1},
Shibani Mukerji\SUP{1},
Sahar Zafar\SUP{1},
Nicole Benson\SUP{1},
Lidia Moura\SUP{1},
John Hsu\SUP{1},
Bradley T. Hyman\SUP{1},
Michael B. Westover\SUP{1},
Deborah Blacker\SUP{1},
Sudeshna Das\SUP{1}
\centering \Email{
\\[\bigskipamount] 
\SUP{1}\{zhong1, 
cmagdamo,
ysheu,
pmohite,
anoori1,
emye,
wendong.ge,
hsun8,
lnbrenner,
grobbins,
smukerji,
sfzafar,
nbenson,
lidia.moura,
john.hsu,
bhyman,
mwestover,
dblacker,
sdas5\}
@mgh.harvard.edu}
\begin{center}\addr Massachusetts General Hospital, Boston, MA\end{center}
}

\begin{document}

\maketitle

\begin{abstract}

 Dementia is under-recognized in the community, under-diagnosed by healthcare professionals, and under-coded in claims data. Information on cognitive dysfunction, however, is often found in unstructured clinician notes within medical records but manual review by experts is time consuming and often prone to errors. Automated mining of these notes presents a potential opportunity to label patients with cognitive concerns who could benefit from an evaluation or be referred to specialist care.  In order to identify patients with cognitive concerns in electronic medical records, we applied natural language processing (NLP) algorithms and compared model performance to a baseline model that used structured diagnosis codes and medication data only. An attention-based deep learning model outperformed the baseline model and other simpler models.
\end{abstract}

\begin{keywords}
NLP, EHR, Dementia
\end{keywords}

\section{Introduction}
\label{sec:intro}
Early detection of dementia is important for improving clinical outcomes and management of dementia, as well as for lifestyle, financial, and future planning for patients and their caregivers \citep{robinson_dementia_2015, borson_improving_2013}. Yet, dementia is not formally diagnosed or coded in claims for over 50\% of older adults living with probable dementia \citep{amjad_underdiagnosis_2018,alzheimers_association_2019_2019}. Tools that screen medical records for warning signs and present the digested information to providers may prove to be an important step for early intervention. 
In this study, we aim to use NLP to detect signs of cognitive dysfunction from clinician notes in electronic health records (EHR) by applying deep learning techniques that have not been hitherto applied to this problem. We present an attention-based transformer model that allows for long text sequences to reveal signs of cognitive concerns and compare its performance to baseline models. 

\section{Related Work}
\label{sec:relatedwork}
Prior works have used pattern based text-analytics to detect dementia in electronic health records. Dementia detection using NLP on provider notes has been applied to a cohort of acute-care EHRs of hip and stroke fracture patients \citep{gilmore-bykovskyi_unstructured_2018}; a sample of patients enrolled in the UCLA Alzheimer's and Dementia Care (ADC) program \citep{reuben_automated_2017}; and a group of patients who had a formal cognitive evaluation by the Mayo Clinic Study of Aging and were hospitalized at their institution \citep{amra_derivation_2017}. These studies demonstrated that the incorporation of NLP on EHR notes improves sensitivity of dementia detection. The current work uses deep learning based NLP, which has achieved numerous breakthroughs when applied to general text thanks to the use of word embeddings and attention-based models \citep{vaswani2017attention,mikolov2013distributed,pennington2014glove,peters2018deep, devlin2018bert}, while it has had limited application on healthcare data. Our work builds upon the idea in \citep{sheu2019can}, where a deep-learning based model was applied to classify the outcome for treatment of depression, a phenotype also not captured well by structured data or term-based NLP, therefore motivating the use of deep learning techniques.

\section{Datasets and Preprocessing}
\label{sec:dataset_preprocessing}

\paragraph{Datasets}
\label{sec:dataset}Our gold-standard dataset consisted of 1K randomly sampled patients at Mass General Brigham (MGB) HealthCare (formerly Partner’s Healthcare, comprising two major academic hospitals, community hospitals, and community health centers in the Boston area). Each patient’s EHR record between 01/01/2018 – 12/31/2018 was reviewed by an expert clinician (neurologist, psychiatrist, or geriatric psychiatrist) to label patients with presence or absence of any cognitive concerns (signs of dementia).

\paragraph{Data Preprocessing}
\label{sec:preprocessing}

For each patient in the gold-standard dataset, we extracted the diagnosis, medication history, and progress notes from the window between 01/01/2018 – 12/31/2018. We flagged all medications related to cognitive concern (Galantamine, Donepezil, Rivastigmine, Memantine) as well as all International Classification of Diseases (ICD) diagnoses pertaining to cognitive concern (ICD 9 codes: 290.X, 294.X, 331.X, 780.93; ICD 10 codes: G30.X and G31.X). For each patient, we added the number of records pertaining to the medications of interest and the number of ICD codes of interest, yielding two features designed to capture the presence of cognitive concerns in the structured data. We filtered for patients that had progress notes. Several preprocessing steps were involved to trim the notes to reduce noise, and note length was decreased by 50\% after these steps, details of which are described in \appendixref{apd:app6}. The preprocessed notes were used as input for all three models that use notes as input. For the model based on regular expression matches, we constructed 15 categories of keywords (\appendixref{apd:app1}). Each category contained multiple regular expressions that were then flagged within our filtered set of notes. We then added the number of pattern matches per patient per category across all notes, resulting in 15 features per patient.

The final dataset of 767 patients was randomly split between train (90\%) and hold-out test (10\%) sets. Validation datasets were split within the train set using different methods for the various models as described in the Methodology section. \tableref{tab:example-booktabs} shows the demographics of the patients in the training and test sets.

\begin{table*}[hbtp]
\floatconts
  {tab:example-booktabs}
  {\caption{Demographics of Train/Val and Test Sets}}
 {
  \begin{tabular}{lccc}
  \toprule
  \bfseries Dataset & \bfseries Age (Mean and SD) & \bfseries \% Female & \bfseries Cognitive Concern (N) \\
  \midrule
  Train/Val (N = 690) & 81.2 (7.4) & 58.2\% & 308 (44.64\%) \\
  Test (N = 77) & 80.19 (6.79) & 54.5\% & 34 (44.16\%) \\
  \bottomrule
  \end{tabular}}
\end{table*}

\section{Methodology}
\label{sec:methodology}
We built three models and compared the performance to the baseline  model.
\paragraph{(1) Baseline Model}In model 1, we performed a regularized logistic regression where the cognitive concern label was regressed on counts of medications and ICD codes related to cognitive concern. We used the glmnet package \citep{glmnet} in R and performed L1 regularization. The lambda value in our regularization penalty was selected via 10-fold cross validation (CV), with the optimal lambda chosen to maximize the average area under the Receiver Operator Characteristic (ROC) curve across folds.

\paragraph{(2) Logistic Regression with Regular Expression Counts} In model 2, we again utilized regularized logistic regression. We regressed the cognitive concern label against the 15 features corresponding to counts of the 15 regular expression categories per patient. We similarly used L1 regularization and 10-fold CV to choose our lambda value. 

\paragraph{(3) Logistic Regression with TF-IDF Vectors} In model 3, we performed a TF-IDF (term frequency–inverse document frequency) vectorization to select features based on the term's correlation coefficient, and performed a regularized logistic regression using the cognitive concern label as the outcome. We used different correlation coefficients as thresholds to select word features, and iterated with different lambda values on a validation dataset (10\% from train dataset) to determine our best lambda value and the specific correlation cutoff threshold.  

\paragraph{(4) Transformer Based Sequence Classification Language Model}

In model 4, we utilized a pre-trained language model with an attention mechanism, Longformer \citep{beltagy2020longformer}, which deploys a modified self-attention mechanism whose complexity scales linearly with sequence length and thus allows accommodation of longer sequences. The model was initialized with pre-trained parameters and was later fine-tuned with our labeled training dataset. Hyperparameters were tuned on a held out validation dataset (10\% from the train dataset) for the cognitive concern classification task. 

\section{Experiments and Results}
\label{sec:experiments}

\subsection{Description of Computational Experiments}
 In model 4 with Longformer (model diagram is shown in \appendixref{apd:app4}), we utilized the implementation in the Huggingface Transformer \citep{Wolf2019HuggingFacesTS} and Simpletransformers packages \citep{simple2020thilina}. After text preprocessing, input texts were tokenized with the default tokenizer and converted to embeddings. Due to the limitation of sequence lengths of Longformer, we used the “sliding windows” option available in the Simpletransformers package to automatically slice the input text into windows of 4096 tokens, with 20\% overlapping content between windows \emph{i} and \emph{i+1}. We used the Adam optimizer \citep{kingma2014adam} and tuned the model with an initial learning rate of 7.09e-6, Epsilon value of 1e-9, and 4 training epochs. Classification of each window was aggregated (mode) to obtain patient level class assignments. All performance metrics below are based on patient level class assignments.  

\begin{table*}[hbtp]
\floatconts
  {tab:performance}
  {\caption{Model Performance Comparison}}
  {\begin{tabular}{lllllllll}
  \toprule
  \bfseries Model &\bfseries AUC& \bfseries Accuracy & \bfseries FP & \bfseries FN &\bfseries Sensitivity &\bfseries Specificity &\bfseries PPV &\bfseries NPV\\
  \midrule
  Model 1 & 0.79 & 0.82 & 0 & 14 & 0.59 & 1.00 & 1.00 & 0.75\\
  Model 2 & 0.88 & 0.84 & 4 & 8 & 0.76 & 0.91 & 0.87 & 0.83\\
  Model 3 & 0.90 & 0.86 & 6 & 5 & 0.85 & 0.86 & 0.83 & 0.88\\
  Model 4 & 0.93 & 0.90 & 1 & 7 & 0.79 & 0.98 & 0.96 & 0.86\\
  \bottomrule
  \end{tabular}}
\end{table*}

\subsection{Model Results and Comparison}
 We evaluated each model using a prediction threshold to maximize the accuracy and calculated performance metrics based on that threshold. The performance of the four models are shown in \tableref{tab:performance}.  Specifically, PPV, the positive predictive value is defined as the percentage of true positives among all cases predicted as positive; NPV, the negative predictive value is defined as the percentage of true negatives among all cases predicted as negative. The baseline model had an AUC-ROC of 0.79 and while the specificity was 1.0, the sensitivity was 0.59. There were 14 false negatives; of these patients with no record of ICD codes or medications in EHR, 8 were patients with subjective concern or mild cognitive impairment (MCI) and 6 were patients with dementia. The regular-expression based logistic regression had a substantial improvement in performance over the baseline model with an AUC-ROC of 0.88. The percentage of patients with dementia-related concepts shown in \appendixref{apd:app1}; terms related to memory and dementia-related medications had the highest coefficients in the regression model. The sensitivity of detection improved to 0.76 but the specificity was 0.91. The TF-IDF model demonstrated higher performance than the regular expression-based model (AUC-ROC=0.90). The 20 words with the highest correlation coefficients using TF-IDF vectorization are shown in  \tableref{table:tdf} in the Appendix \ref{apd:app2}.  The deep learning model outperformed all the models (AUC-ROC=0.93). The patients misclassified in the models were different, demonstrating that each model had its own advantages.

\section{Conclusion and Future Work}
\label{sec:conclusion}

We applied NLP algorithms to identify patients with cognitive concerns in EHR and compared model performance. While the deep learning model's performance was the best, it was only marginally better than the term based NLP models. We posit that deeper representations will be required for more complex tasks requiring syntactical and contextual information such as classifying the stage of cognitive impairment: MCI, mild, moderate or severe dementia. Our gold standard set had a relatively smaller proportion of patients with subjective concerns or mild cognitive impairment (28.6 \% with cognitive concern), and the overall sample size was small. To address these issues, we plan to implement an active learning loop, starting from querying additional at-risk patients over age 65 and without a dementia related ICD code or medication, and apply the fine-tuned model to derive the probability of having cognitive concerns for these patients. For the edge cases, the notes will be manually reviewed and labeled. To improve the efficiency of this review process, we designed an annotation tool that highlights the sections with regular expression matches or higher attention weights (UI design in Appendix \ref{apd:app5}). The new gold-standard data will serve as the basis for the next iteration of the active learning loop to further improve model performance and potentially detect patients with an earlier stage of cognitive decline.

\clearpage

\bibliographystyle{plain}
\bibliography{jmlr-sample}


\clearpage
\appendix

\section{Clinical Note Data Preprocessing Steps}\label{apd:app6}

In order to reduce the input unstructured data dimensionality and work with a smaller vocabulary, we adopt a series of preprocessing steps  to trim the notes to reduce noise for all three proposed models. Those steps include:
\begin{enumerate}

 \item Removed all empty lines and multiple white spaces.
 \item Removed the fixed header section which only contains non-diagnostic info. \item Remove all note section content that was verified to be unimportant by experts (either not relevant to the cognitive concern detection or contained the same info already stored in structured data format outside the note space).
 \item Utilized SpaCy's default name entity recognition package to detect and remove certain non-relevant name entities includes date, time, person, quantity info.
 \item Removed all special characters and all other numbers after SpaCy entity removal.
 \item Normalized all rest of the text to lower case.
 \item For model 4 only, we further used the tokenizer provided by the pre-trained Longformer \citep{beltagy2020longformer} model as tokenization in the model preprocessing steps, since the Longformer tokenizer, like other attention-based model, has a its particular way of handling out-of-vocabulary words.
\end{enumerate}


\hfill \break

\section{TF-IDF Feature Correlation Rank}\label{apd:app2}

\begin{table}[hbtp]
\floatconts
  {tab:tfidf}
  {\begin{tabular}{rrr}
  \toprule 
  \bfseries Rank & \bfseries Terms & \bfseries CorrCoef\\
  \midrule
  1 & dementia    & 0.4364 \\
  2 & memory      & 0.3587 \\
  3 & daughter    & 0.2959 \\
  4 & cognitive   & 0.2955 \\
  5 & alzheimer   & 0.2940 \\
  6 & accompanied & 0.2890 \\
  7 & behavioral  & 0.2809 \\
  8 & unable      & 0.2756 \\
  9 & confused    & 0.2754 \\
  10 & donepezil  & 0.2738 \\
  11 & mental     & 0.2686 \\
  12 & aricept    & 0.2664 \\
  13 & care       & 0.2580 \\
  14 & impairment & 0.2529 \\
  15 & nursing    & 0.2402 \\
  16 & assistance & 0.2365 \\
  17 & nurse      & 0.2306 \\
  18 & living     & 0.2294 \\
  19 & rn         & 0.2285 \\
  20 & dnr        & 0.2271 \\
  \bottomrule
  \end{tabular}
  \label{table:tdf}}
  {\caption{TF-IDF Vector Features Correlation}}
\end{table}

  

\clearpage

\section{Dementia Related Concepts}\label{apd:app1}

\begin{figure}[htbp]
\floatconts
  {fig:regex}
  {\caption{Dementia Related Concepts}}
  {\includegraphics[width=2\linewidth]{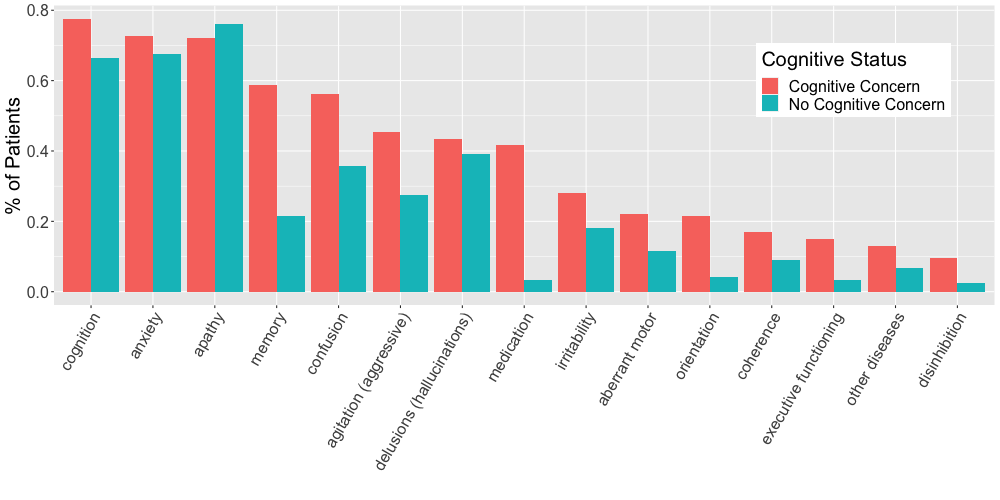}}
\end{figure}
\clearpage

\section{Longformer with Additional Sliding Windows}\label{apd:app4}

\begin{figure}[htbp]
\floatconts
  {fig:longformer}
  {\caption{Longformer Model with Additional Sliding Window Strategy}}
  {\includegraphics[width=2\linewidth]{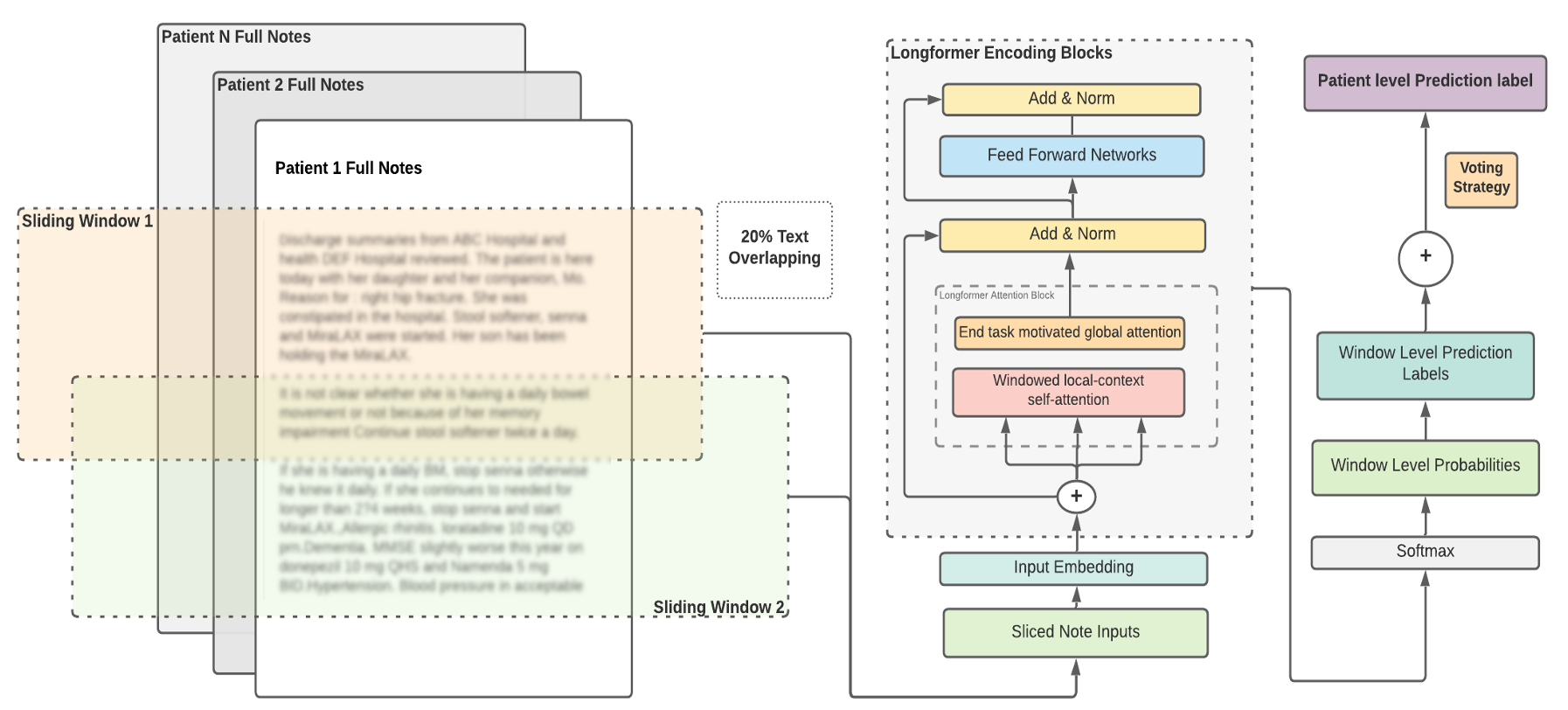}}
\end{figure}
\clearpage

\section{Active Learning Annotation User Interface}\label{apd:app5}

\begin{figure}[htbp]
\floatconts
  {fig:LabelingUI}
  {\caption{Active Learning Annotation User Interface}}
  {\includegraphics[width=2\linewidth]{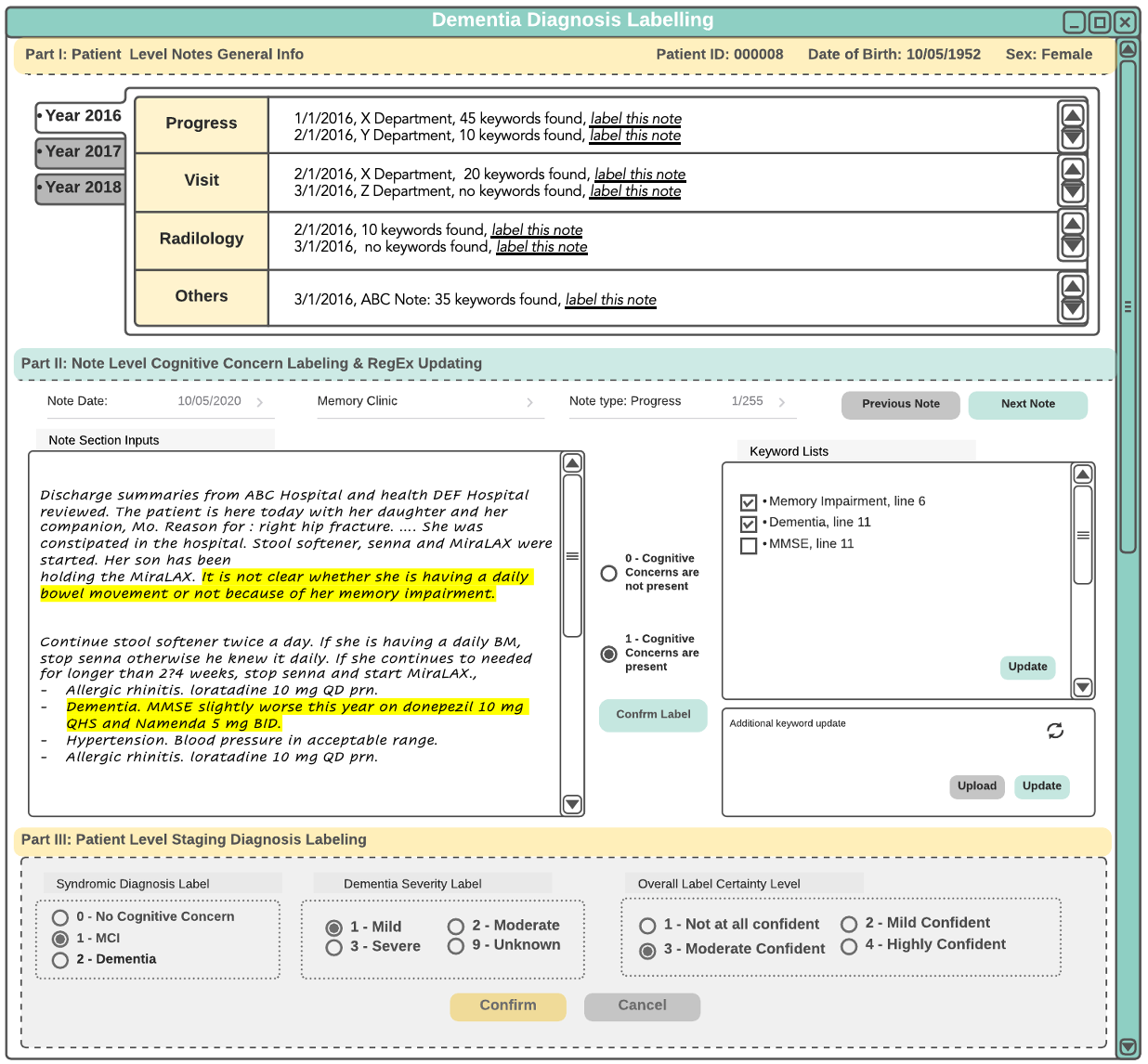}}
\end{figure}

\end{document}